# A Brief History of Learning Classifier Systems:

# From CS-1 to XCS

Larry Bull

Department of Computer Science & Creative Technologies

University of the West of England

Bristol BS16 1QY U.K.

larry.bull@uwe.ac.uk

**Abstract**

The legacy of Wilson's XCS is that modern Learning Classifier Systems can be characterized by their use of rule accuracy as the utility metric for the search algorithm(s) discovering useful rules. Such searching typically takes place within the restricted space of co-active rules for efficiency. This paper gives an historical overview of the evolution of such systems up to XCS, and then some of the subsequent developments of Wilson's algorithm to different types of learning.

## 1. Introduction

Learning Classifier Systems (LCS) are rule-based systems, where the rules are usually in the traditional production system form of "IF condition THEN assertion". An evolutionary algorithm and/or other heuristics are used to search the space of possible rules, whilst another learning process is used to assign utility to existing rules, thereby guiding the search for better rules. The LCS formalism was introduced by John Holland [1976] and based around his more well-known invention – the Genetic Algorithm (GA)[Holland, 1975]. A few years later, in collaboration with Judith Reitman, he presented the first implementation of an LCS in "Cognitive System Level 1" (CS-1) [Holland & Reitman, 1978]. Holland then revised the framework to define what would become the standard system for many years [Holland, 1980]. However, Holland's full system was somewhat complex and practical experience found it difficult to realize the envisaged behaviour/performance, despite numerous modifications (e.g., see [Wilson & Goldberg, 1989]), and interest waned. Some years later, Stewart Wilson introduced a form of LCS in which rule fitness is calculated solely by the accuracy of the predicted consequences of rule assertions – the "eXtended Classifier System" (XCS) [Wilson, 1995]. The following two decades have seen a resurgence in the use of LCS as XCS in particular has been found able to solve a number of well-known problems optimally (e.g., see [Butz, 2006]). Perhaps more importantly, LCS have been applied to a number of real-world problems (e.g., see [Bull, 2004]), particularly data mining (e.g., see [Bull et al., 2008]), to great effect. Formal understanding of LCS has also increased (e.g., see [Bull & Kovacs, 2005]). The purpose of this paper is to provide some historical context to the area of modern accuracy-based Learning Classifier Systems before presenting some of the main developments since the introduction of XCS twenty years ago (see [Lanzi & Riolo, 2000] for a previous historical review). The use of evolutionary algorithms to design whole rule sets, i.e., so-called Pittsburgh-style LCS [Smith, 1980], is not considered here.

## 2. The Evolution of LCS

Holland developed the LCS formalism as an approach to reinforcement learning, that is, learning through trial-and-error. Reinforcement learning methods seek to ascertain the value of executing each possible action (assertion) available within each state (condition) of a given problem (see [Sutton & Barto, 1998] for an introduction). Within psychology, the study of trial-and-error learning can be traced back to Edward Thorndike and his "Law of Effect" [Thorndike, 1911], and within computer science to Alan Turing and his "P-type unorganised machine" [Turing, 1948]. Whilst Farley and Clark [1954] were perhaps the first to implement reinforcement learning within a computer, Holland was influenced by Arthur Samuel's seminal early work on draughts/checkers [Samuel, 1959], which itself drew upon Claude Shannon's work on chess, seemingly the first consideration of learning a value function through experience [Shannon, 1950]. Samuel [1959] described a scheme for adjusting temporally successive estimates of the end reward value from a sequence of moves (improved in [Samuel, 1967]).

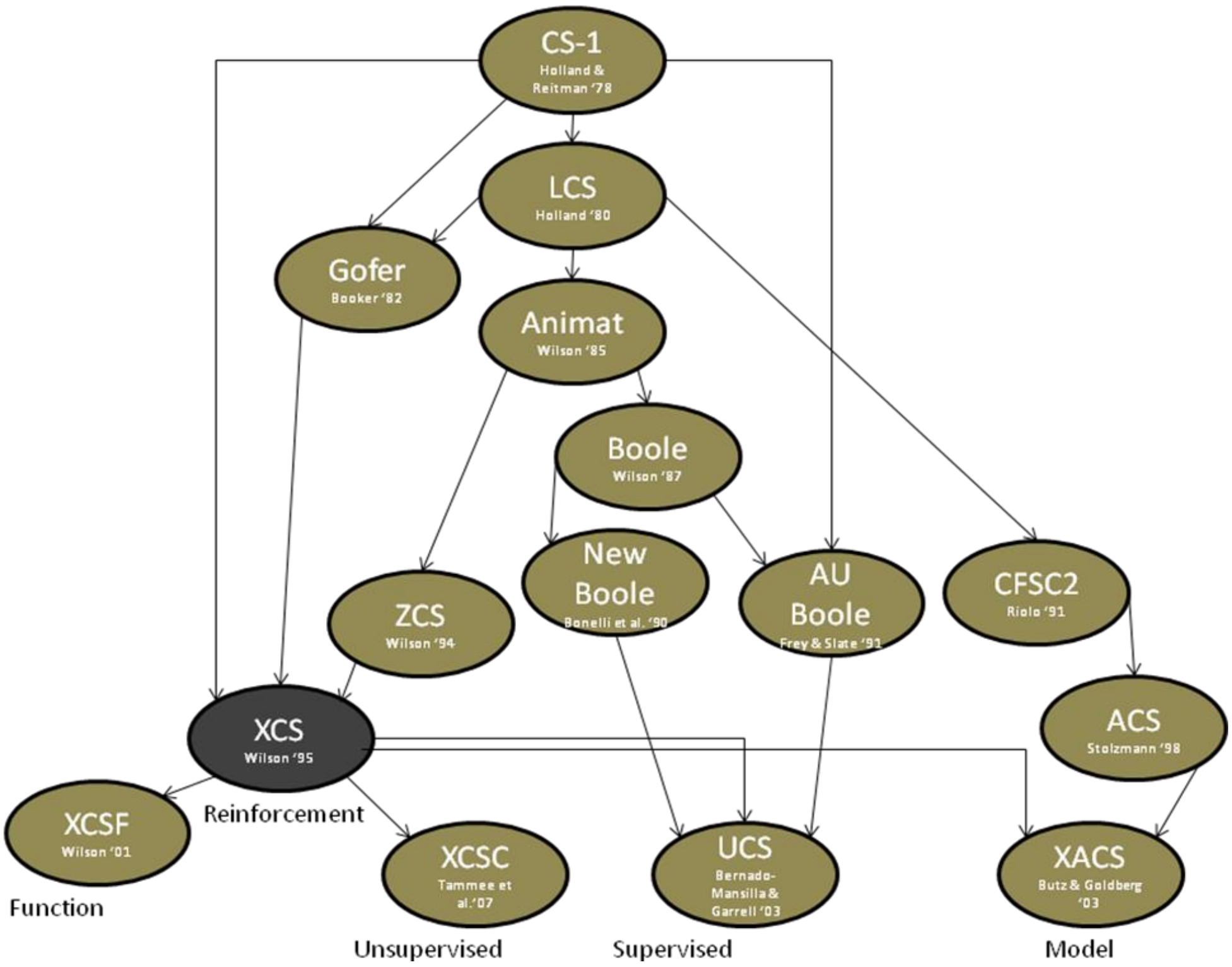

**Figure 1**: Learning Classifier Systems' family tree.

Holland's [1976] interest was in how an artificial system may continuously adapt to novelty, significantly, extending previous studies by also considering how to build suitable knowledge representations thereby enabling flexible, continual learning through trial-and-error. A variant of his Genetic Algorithm was incorporated as an effective approach to this ability. The suggestion that a simulated evolutionary process may prove useful in artificial systems was first made by Turing [1948], with early implementations within a computer by Fraser [1957] and Box [1957] (see [Eiben & Smith, 2003] for an introduction). The combined reinforcement learning and evolutionary computing architecture was termed CS-1 [Holland & Reitman, 1978].

Figure 1 shows a family tree of the LCS considered as the significant steps in the evolution of XCS from CS-1 to be discussed in this paper, beginning with CS-1.

## 2.1 Cognitive System Level 1

On each discrete time cycle, CS-1 receives a binary encoded description of the current state of its environment. The system determines an appropriate response based on this input, its last action, and the current contents of an internal memory space, termed a message list (Figure 2). The rule-base consists of a population of $N$ condition-assertion rules or "classifiers". The rule conditions are strings of characters from the ternary alphabet {0,1,#}. The # acts as a wildcard allowing generalisation such that the rule condition 1#1 matches both the input 111 and the input 101, for example. Rule assertions contain both an action and an internal message, both formed from binary strings. All rule components are initialised randomly. Also associated with each rule are a number of parameters, including age, frequency of use, and a prediction of the typical reward received from its use, which is also the fitness metric (explained later).

On receipt of an input message, the rule-base is scanned and any rule whose condition matches the external message, the content of the message list, and the previous action becomes a member of the current "match set" [M]. A heuristic considering aspects such as the specificity of matching and the predicted future reward is then used to determine the top ten eligible rules. The system response is then chosen probabilistically from that set. The chosen rule’s action is executed in the environment, the message list updated, its age halved, and frequency increased.

CS-1 uses an epochal reinforcement learning scheme such that the identification of all rules that have provided an action is recorded, in order, between rewards. If an external reward is received, the predicted reward of each rule in this set [E] is adjusted at a rate inversely proportional to their frequency parameter “to reflect their accuracy in anticipating this reward. Those predicted payoffs that were consistent with (not greater than) this reward are maintained or increased; those that overpredicted are significantly reduced” [Holland & Reitman, 1978]. A further heuristic is applied to the predictions such that the actual value of reward used to update each member of [E] is “attenuated”, an adjustment based on the relative size of reward predicted by rules and by their successors in [E], it being incremented each time the latter is higher than the former. “This parameter is highly correlated with the delay between a response and the reward" [Holland & Reitman, 1978] and may be seen as an early form of temporal difference learning.

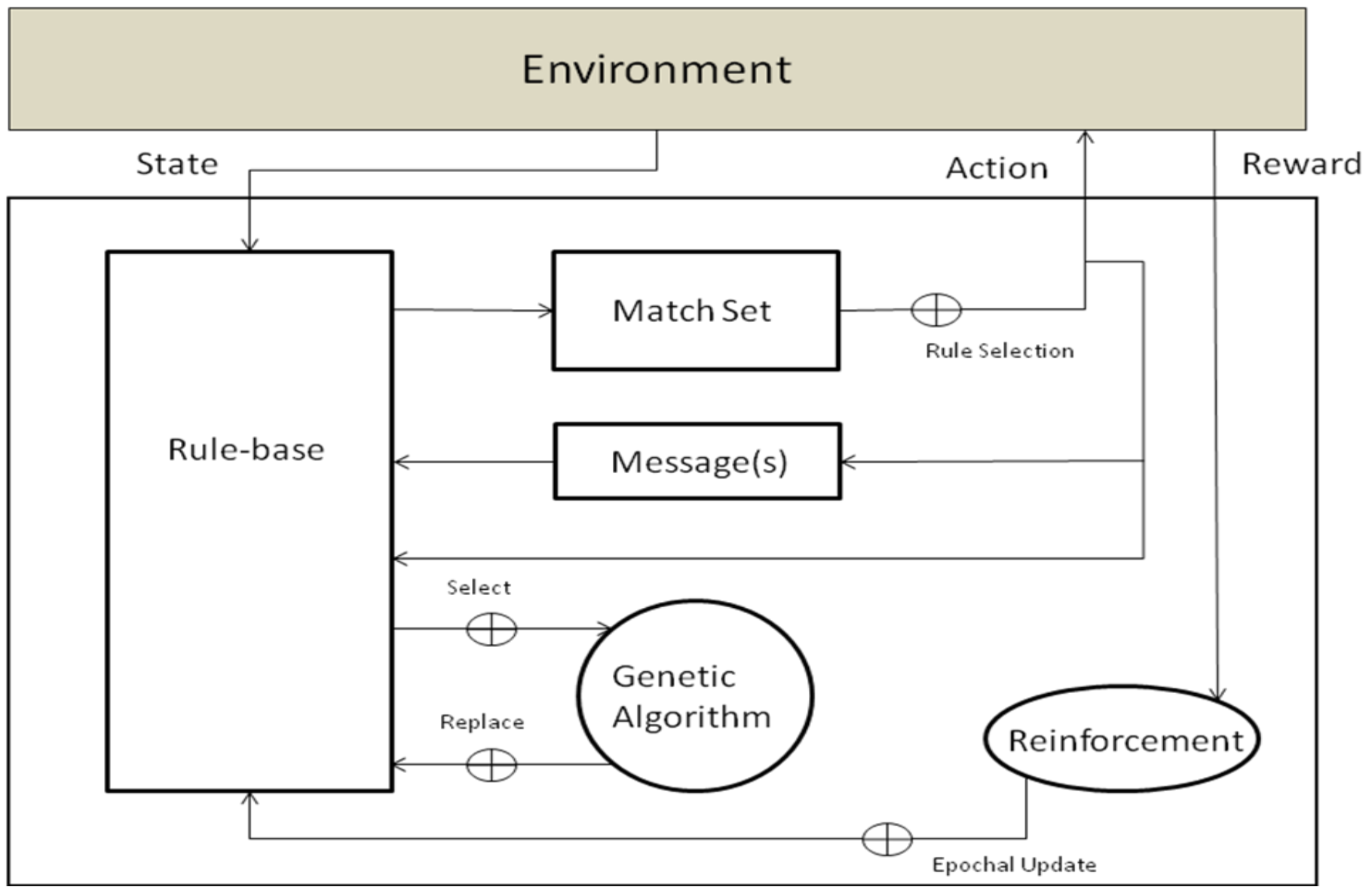

**Figure 2**: Schematic of CS-1.

After every ten rewards received, the contents of the rule-base are altered by the simulated evolutionary process of the GA. The implemented CS-1 could take one of two actions and so ran the evolutionary search process within the two sub-populations, that is, action niches. Fitness proportionate selection using the predicted reward of each rule as the fitness value picks two parents from the rule-base population. These are then combined using one-point crossover (mutation is not mentioned). One of the two offspring produced is selected at random to be inserted into the rule-base. Replacement uses the age of rules. "Recall that a classifier with a poor predicted payoff rarely wins competitions; without a win, its age increases steadily. Age therefore, reflects the classifier's quality as well as its frequency of use. To make room for the new classifier therefore, one with an old age is deleted." [Holland & Reitman, 1978]. Moreover, from the set of oldest rules within the niche, the one closest in Hamming(-like) distance is chosen; a form of crowding is used.

CS-1 was shown able to solve a simple maze task with seven locations, two actions, and two types of reward,

before being applied to an extended maze. Holland and Reitman report faster learning of the second maze using a system previously trained on the smaller maze, in comparison to a naïve system. Analysis of the external input patterns indicates minor changes in effective general rules in the smaller maze are close in rule-space to those in the larger, as might be expected (see [Iqbal et al., 2014] for related recent work). Wilson [1981] was also able to use a CS-1 based system for a camera centring control problem, which included some complex image processing to create the binary inputs to the LCS.

LCS aim to build an efficient representation of any underlying regularities within the given problem domain during learning. The inherent pressure within CS-1 to discover *maximally general* rules - rules which aggregate the most problem states together from which the same action results in the same reward - over more specific (less # symbols in their condition) but equally accurate predictors comes from the evolutionary deletion scheme. More specific rules tend not to match so often and so their age increases more rapidly than more general, but also accurate rules; the probability of removal of specific rules from the rule-base increases with specificity. Similarly, the pressure to remove *over general* rules – rules that aggregate too many states together such that the level of reward received from their use varies – comes from the reinforcement scheme and evolutionary selection. Over or under prediction of a reward value results in a significant reduction in the predicted reward of a rule, the parameter used as the fitness measure for reproduction by the GA; inaccurate rules have lower fitness and hence are unlikely to be selected.

However, the described system struggles to maintain more than one or a very few rules within a population. That is, the GA tends to converge upon a single (maximally general) solution. This explains why CS-1 runs the GA in the two explicit action sub-population niches. In the two mazes, CS-1 always started in the middle and had to maintain the same action across a number of states to an end goal state where reward was given. Whether the system should go left or right depended upon an internal value. Hence a rule which generalised over all the states to the left and one which generalised over all the states to the right was the optimal solution. By running the GA in two niches based on actions, both were sustainable indefinitely.

With this apparent limitation, Holland subsequently altered a number of aspects of CS-1. Importantly, Wilson would return to the use of reward prediction accuracy and frequency of use in XCS.

## 2.2 Holland’s Standard Architecture

A few years later Holland [1980] revised CS-1 and described what would become the standard architecture, here termed “Learning Classifier System” (LCS). It should be noted that Holland seems not to have used the prefix “learning” at the time; Goldberg [1985] may have been first to add the emphasis. The main change from CS-1 was to introduce a reinforcement learning scheme based upon an economic metaphor, known as the “bucket brigade” (after the water passing chains of fire fighters), in which rule utility is judged by the accruement of credit. In this way, rules acting in temporal chains leading to external reward are viewed as the middlemen of supply and demand chains. Rules maintain a single parameter of credit (termed strength) received. This is used both for action selection and in rule discovery by the GA. The message list is extended to enable multiple rules to post their assertions. Rule conditions no longer have a fixed structure to consider the current environmental state, the contents of the message list and the last action. Instead, all conditions and assertions are of the same length, with conditions also able to include a logical NOT. Assertions are now built from the same alphabet as conditions {0,1,#} such that information may “pass through” from either the condition or the string (external input or internal message) which the rule matches where a # exists.

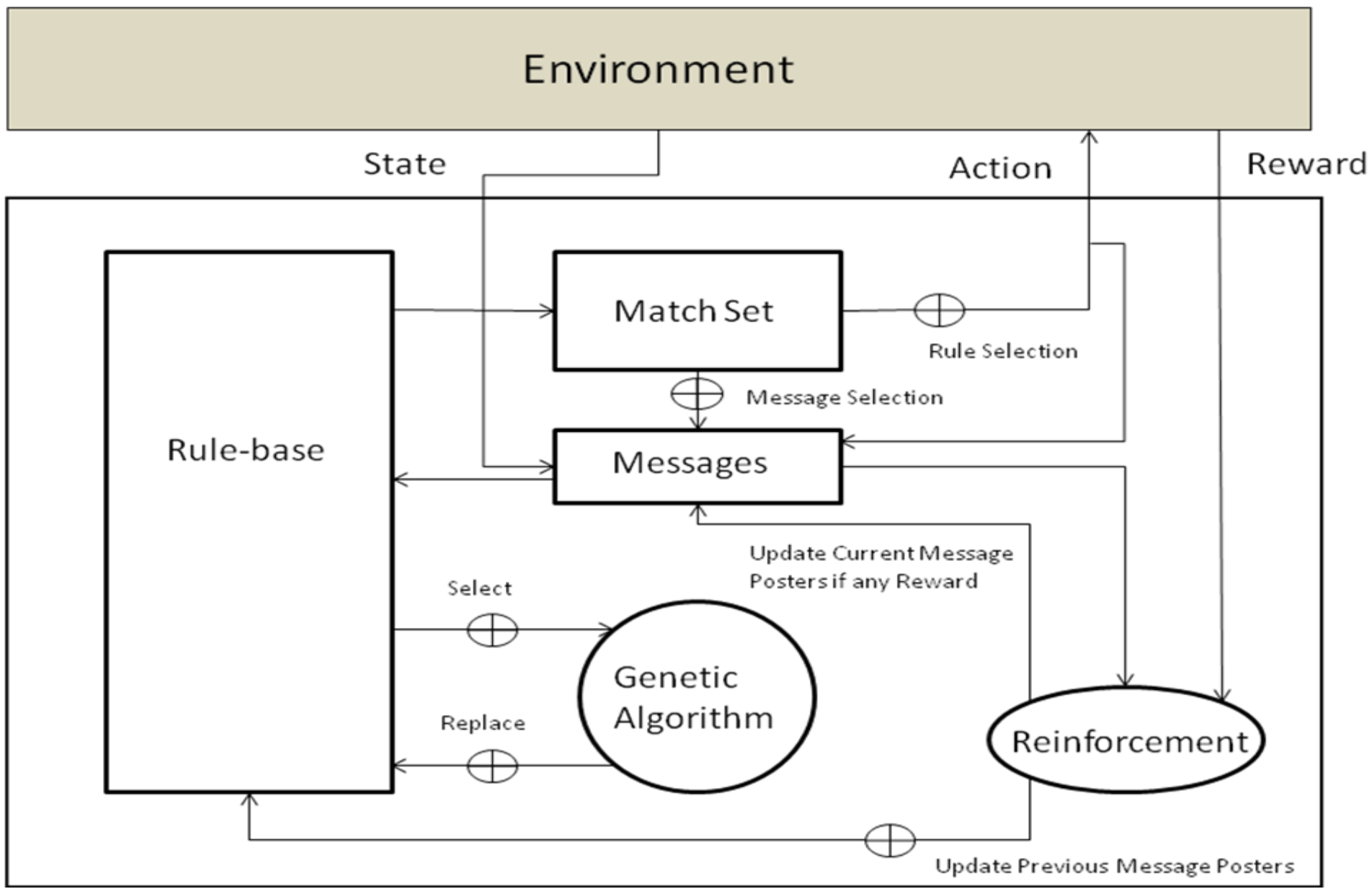

**Figure 3**: Schematic of Holland’s LCS.

On each cycle, a binary external state description is placed onto the message list, the rule-base is scanned, and any rule whose condition matches the external message and/or the other contents of the message list becomes a member of [M] (Figure 3). Rules are selected from those comprising [M], through a bidding mechanism, firstly to become the system's external action and then to post their assertion onto the (fixed size) message list for the next cycle. This selection is performed by the roulette wheel scheme based on rule bids. Rules' bids consist of two components, their strength and the specificity (fraction of non-# bits) of their condition. Further, a constant (here termed $\beta$, where $0<\beta<1$) is factored in, i.e., for a rule $C$ in [M] at time $t$:

$$\text{Bid}(C,t) = \beta.\text{specificity}(C).\text{strength}(C,t)$$

Reinforcement consists of redistributing bids made between subsequently chosen rules. The bid of each winner at each time-step is placed in a "bucket". A record is kept of the winners on the previous time step and they each receive an equal share of the contents of the current bucket; fitness is shared amongst concurrently activated rules. If a reward is received from the environment then this is paid to the winning rule which produced the last system output. Although, whether all rules that have posted a message share the external reward appears to vary in the literature, being both included [Holland, 1985] and excluded [Holland, 1986]. "Thus, the bucket brigade assures that early-acting, stage-setting classifiers receive credit if they (on average) make possible later, overtly rewarding acts" [Holland, 1986]. The emphasis upon average ability relaxes the previous explicit focus on accurately predicting reward; there is an apparent reduction in the selective pressure for consistent behaviour which formed the basis of CS-1. With hindsight, this change was perhaps the most significant between the two LCS.

As noted above, the periodically applied GA uses rule strength to select two parents, these are then combined using one-point crossover and mutated. Both offspring are inserted into the rule-base, replacing rules chosen inversely proportional to their strength. Since reward is shared amongst rules, the GA is in principal able to maintain multiple useful rules within the rule-base (discussed later).

A number of other mechanisms were proposed by Holland but for the sake of clarity they are not described in detail here. In particular, the idea that hierarchical rule associations could emerge via specific rules out-bidding more general rules in certain important situations, and extra “tag” regions of conditions and assertions being added would aid the formation of sequential induction (see [Holland et al., 1986] for a full treatment). These ideas do not feature in modern LCS (see [Smith et al., 2010] for an exception).

## 2.3 GOFER

Booker [1982] presented a form of Holland’s standard LCS which extends the principle of using a GA to discover any underlying regularities in the problem space, dividing the task of learning such structure from that of supplying appropriate actions to receive external reward (see [Booker, 1988] for an overview]). Here a separate LCS exists for each of these two aspects. A first LCS receives binary encoded descriptions of the external environment, with the objective to learn appropriate regularities through generalizations over the “perception” space. This is seen as analogous to learning to represent categories of objects. The matching rules not only post their messages onto their own message list but some are passed as inputs to a second LCS. The second LCS therefore only receives reward when it correctly exploits such categorizations with respect to the current task. GOFER contains a number of innovations including partial matching and rule excitation levels, however it is the use of restricting the actions of rule-discovery to concurrently active rules which has proven most influential (see [Booker, 1985]). Here parents are chosen from within a given [M] thereby avoiding the mixing of rules with generalizations which (potentially) consider markedly different aspects of the problem. Booker [1989] later extended the idea to trigger the GA during learning whilst also leaving it running at a constant rate under the reinforcement process as Holland did. In particular, rules maintain an approximation of their “consistency”, a measure of the variance in the reward they receive. If a given percentage of rules in [M] have a level of inconsistency above a threshold, the fitness of all consistent rules is increased and the GA run: “ … consistent classifiers are thereby made more attractive to the genetic algorithm” [ibid.]. XCS uses both a form of triggered niche GA and rule consistency.

## 2.4 ANIMAT and ZCS

Having explored CS-1 for a real-world task, Stewart Wilson began to develop versions of Holland’s LCS as an approach to understanding animal/human intelligence through the computer simulation of simple agents in progressively more complex domains - termed the animat approach. The first of these, ANIMAT [Wilson, 1985], makes a number of simplifications to Holland's architecture. In particular, the message list is removed and matching rules are grouped by their action in the bucket brigade process, forming actions sets [A]. The GA is also sensitive to rule actions, somewhat akin to CS-1: a first parent is chosen based upon its strength from the rule-base, the second is then chosen from the subset of the population with the same action. ANIMAT controlled a simple agent in a 2D gridworld, able to sense the contents of the eight locations surrounding it and able to move in each such direction if clear. Wilson showed learning was possible such that effective paths to food reward signals were discovered. However, he noted that the system had "nothing which preferentially reinforces the most expeditious classifiers" [ibid.]. To encourage the shortest path to reward from a given start location, rules were extended to maintain an estimate of the number of subsequent steps to reward from their use, updated locally based upon the estimates of successor rules. This was factored into action selection via dividing strength by distance. ANIMAT also includes a guided recombination operator, replacing dissimilar bits in parent conditions with a # to aid the formation of useful generalizations.

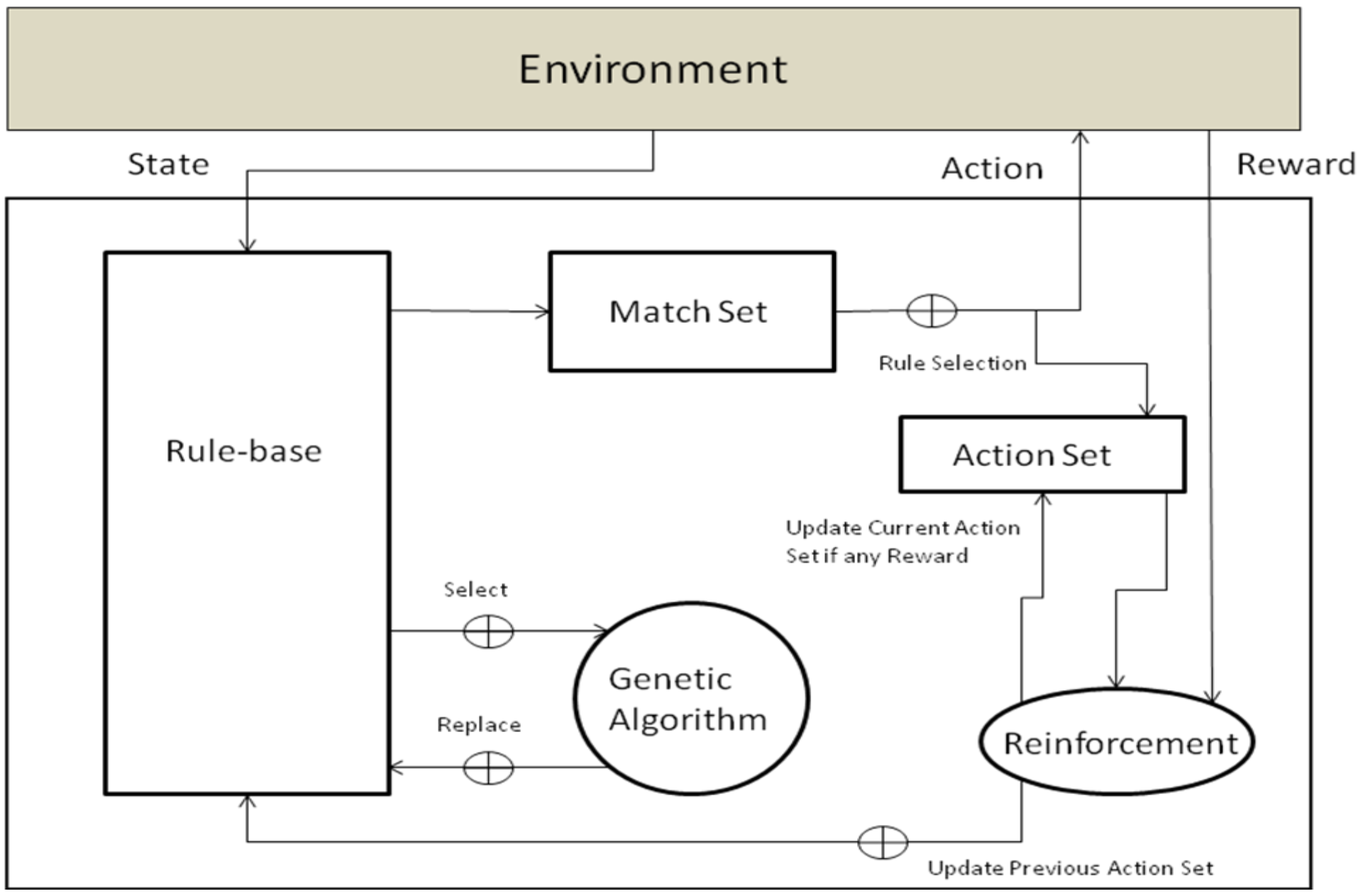

**Figure 4**: Schematic of Wilson's ZCS.

Wilson later returned to ANIMAT, further simplifying it in his "zeroth-level" classifier system (ZCS) [Wilson, 1994] (Figure 4). Importantly, the bucket brigade was again modified to incorporate a mechanism from temporal difference learning [Sutton & Barto, 1981] (see also [Dorigo & Bersini, 1994] for an early connection). Here the fraction of the total strength of a given [A] in the bucket is further reduced by a discount rate $\gamma$ ($0<\gamma<1$) before being shared equally amongst the rules of the previous action set $[A]_{-1}$. Discounting allows systematic control over the influence of future rewards, replacing Wilson's previous distance approximation mechanism. The effective update of action sets is thus ($0<\beta<1$):

$$\text{strength}([A],t+1) = \text{strength}([A],t) + \beta.[\text{ Reward} + \gamma.\text{strength}([A]_{+1}) - \text{strength}([A],t)\ ]$$

To give increased focus to the search, rules in a given [M] but not [A] have their strengths reduced by a tax; rules can only persist if they regularly receive (high) reward. A "create" mechanism in ANIMAT is also retained in ZCS, but slightly modified. Here, if an [M] is empty or if the total strength of [M] is below a given threshold, a new rule is created to cover the current environmental input, randomly augmented with some #, and given a random action. The action niche restriction and generalization mechanisms of the GA are removed. Parental rules give half of their strength to their offspring under the GA which fires at a fixed rate $\rho$.

Results with ZCS indicated it was capable of good, but not optimal, performance [Wilson, 1994][Cliff & Ross, 1995]. Wilson [1994] also included a version of the off-policy temporal difference learning algorithm Q-learning [Watkins, 1989] to some benefit. He also proposed to use the triggered niche GA of GOFER on top of the panmictic/global scheme described above. Bull [2005] showed the potential for disruption of the reward sharing scheme using just a niche GA in a similar LCS but no combination is known. It has been shown that ZCS is capable of optimal performance in a number of well-known test problems but that it appears to be particularly sensitive to some of its parameters [Bull & Hurst, 2002], and it has been shown to outperform XCS in classes of noisy domain [Stone & Bull, 2005]. XCS maintains a number of ZCS's basic features but makes significant alterations.

## 2.5 BOOLE, NEWBOOLE and AU-BOOLE

After introducing a number of modifications to Holland's architecture in ANIMAT, Wilson presented a specialised form designed for reinforcement learning tasks where immediate reward is given. In particular, his BOOLE system was designed for binary decision tasks [Wilson, 1987]. BOOLE maintains the [M]→[A] mechanism of ANIMAT, also removing the message list. The GA no longer restricts selection of the second parent to having the same action when using crossover, and reproduction causes the strength of parents to be reduced and donated to offspring akin to the mechanism later used in ZCS. It has been shown that reducing strength can create a pressure for more general rules as they update more frequently and therefore regain reward faster [Bull, 2005]. Again, as in the later ZCS, rules in [M] but not [A] have their strengths reduced by a tax. BOOLE was shown able to learn two- and three-address bit multiplexer problems (6MUX and 11MUX, respectively), with the effects of varying the tax rates, genetic operators and including a reward bias based upon the degree of generalisation in rules explored.

Bonelli et al. [1990] made the significant step of presenting a form of LCS for supervised learning tasks, that is, tasks where the correct response is known at the point of internal updating. Extending BOOLE, they noted that the set of rules in [M] providing the correct response, regardless of whether they formed [A], should receive reward. Hence they split [M] into the correct set [C] and incorrect set Not[C] for their NEWBOOLE system. BOOLE's uses of taxes and a bias in the distribution of reward based upon generality were kept. They showed significant improvement in learning speed compared to BOOLE and to an artificial neural network using backpropagation on the 6MUX and 11MUX tasks. Hartley [1999] showed NEWBOOLE to be competitive with XCS on a well-known set of binary classification tasks, although XCS's maintenance of a full state-action-reward map gave it an advantage in some forms of non-stationary task (see [Bull & Hurst, 2002] for discussion).

Seemingly independently, Frey and Slate [1991] also presented a variant of BOOLE for supervised learning tasks in which they also update the correct set within [M] regardless of the output. Having struggled to find the correct balance of taxing and bid biasing for a letter recognition task, with reference to Holland's [1976] original ideas, they introduced the accuracy-utility system (here termed AU-BOOLE). Here each rule maintains two parameters: accuracy, the ratio of correct bids to total bids made; and, utility, the ratio of correct bids when chosen to total number of times chosen as the output. Accuracy is used in bidding in [M] and for reproduction, and utility is used for deletion. Whilst performance with AU-BOOLE was found to be similar to their version of NEWBOOLE, they report greater ease in finding useful parameters. As will be discussed, these ideas have been incorporated into XCS, resulting in the "sUpervised Classifier System" (UCS) [Bernado Mansilla & Garrell, 2003].

## 2.6 CFSC2 and ACS

Holland and Reitman [1978] suggested a number of extensions to CS-1 at the end of their paper, particularly ways by which to learn more sophisticated cognitive maps than the stimulus-response relations they had achieved. "Cognitive maps allow the system to use lookahead to explore, without overt acts, the consequences of various courses of action." [ibid.]. Again, following Samuel [1959], they describe a scenario of rules being linked over system cycles through the message list which do not cause external actions on each step. Holland [1990] later returned to this aspect, proposing that the aforementioned extra "tag" regions of conditions and assertions that can be added as arbitrary patterns would aid the formation of sequential induction of the necessary form: IF condition AND assertion THEN next-condition. That is, Holland did not seek to change the rule structure from his standard LCS to this direct form. Riolo [1991] was first to implement lookahead capabilities within LCS with his CFSC2. He allowed the system to execute more than one cycle before providing an action, added tags along the lines Holland [1990] had suggested, and introduced an extra strength parameter to represent the predictive accuracy of a rule. Through tags, rules are either connected to external or internal events, or both. Bidding is adjusted to also factor the accuracy of predicted next states of a rule (if any). Rule chains which accurately map features in simple mazes with or without overt reward (latent learning) are reported to emerge under a rule discovery process which is driven by internal and external messages rather than a GA. That is, when no rules match or none are chained across system cycles, various heuristics are used to form appropriate rules via tags. Roberts [1993] presented a related approach within ANIMAT which maintained "followsets", time-stamped information regarding rewards received or next states obtained after a rule had fired. The value of such rewards is factored into rule strengths.

Wilson [1995] proposed altering the rule structure to contain the anticipated next state, with an "expecton" in XCS. Stolzmann [1998] presented a system in which such a rule structure is used (the expecton component termed the "effect") – the Anticipatory Classifier System (ACS). Drawing upon a learning theory from cognitive psychology, sub-populations of rules are learned per [A] via the specialisation of rules based upon the environmental input, both in the condition and effect components. Rule utility is represented by the accuracy of anticipations whilst external reward is used in bidding. A famous experiment with rats in a T-maze [Seward, 1949] is simulated and the results indicate similar behaviour from the ACS. A combination of ACS and XCS has been presented to achieve such model learning, as will be discussed (e.g., see [Butz & Goldberg, 2003]).

## 3. Wilson's XCS

The most significant difference between XCS (Figure 5) and all other LCS prior to its presentation is that rule fitness for the GA is not based on the amount of received by rules in anyway but purely upon the accuracy of predictions ($p$) of reward. The intention in XCS is to form a complete and accurate mapping of the problem space (rather than simply focusing on the higher payoff niches in the environment) through efficient generalizations: XCS learns a value function over the complete state-action space. On each time step a match set is created. A system prediction is then formed for each action in [M] according to a fitness-weighted average of the predictions of rules in each [A]. The system action is then selected either deterministically or randomly (usually 0.5 probability per trial). If [M] is empty covering is used.

Fitness reinforcement in XCS consists of updating three parameters, $\varepsilon$, $p$ and $F$ for each appropriate rule; the fitness is updated according to the relative accuracy of the rule within the set in five steps:

i) Each rule's error is updated: $\varepsilon_j = \varepsilon_j + \beta(\,|\,Reward - p_j\,| - \varepsilon_j)$

ii) Rule predictions are then updated: $p_j = p_j + \beta\,(Reward\text{-}p_j)$

iii) Each rule's accuracy $\kappa_j$ is determined: $\kappa_j = \alpha(\varepsilon_0/\varepsilon)^v$ or $\kappa$=1 where $\varepsilon < \varepsilon_0$ where $\nu$, $\alpha$ and $\varepsilon_0$ are constants controlling the shape of the accuracy function.

iv) A relative accuracy $\kappa_j'$ is determined for each rule by dividing its accuracy by the total of the accuracies in the action set.

v) The relative accuracy is then used to adjust the classifier's fitness $F_j$ using the moyenne adaptive modifee (MAM) procedure: If the fitness has been adjusted 1/β times, $F_j = F_j + \beta(\kappa_j' - F_j)$. Otherwise $F_j$ is set to the average of the values of $\kappa'$ seen so far.

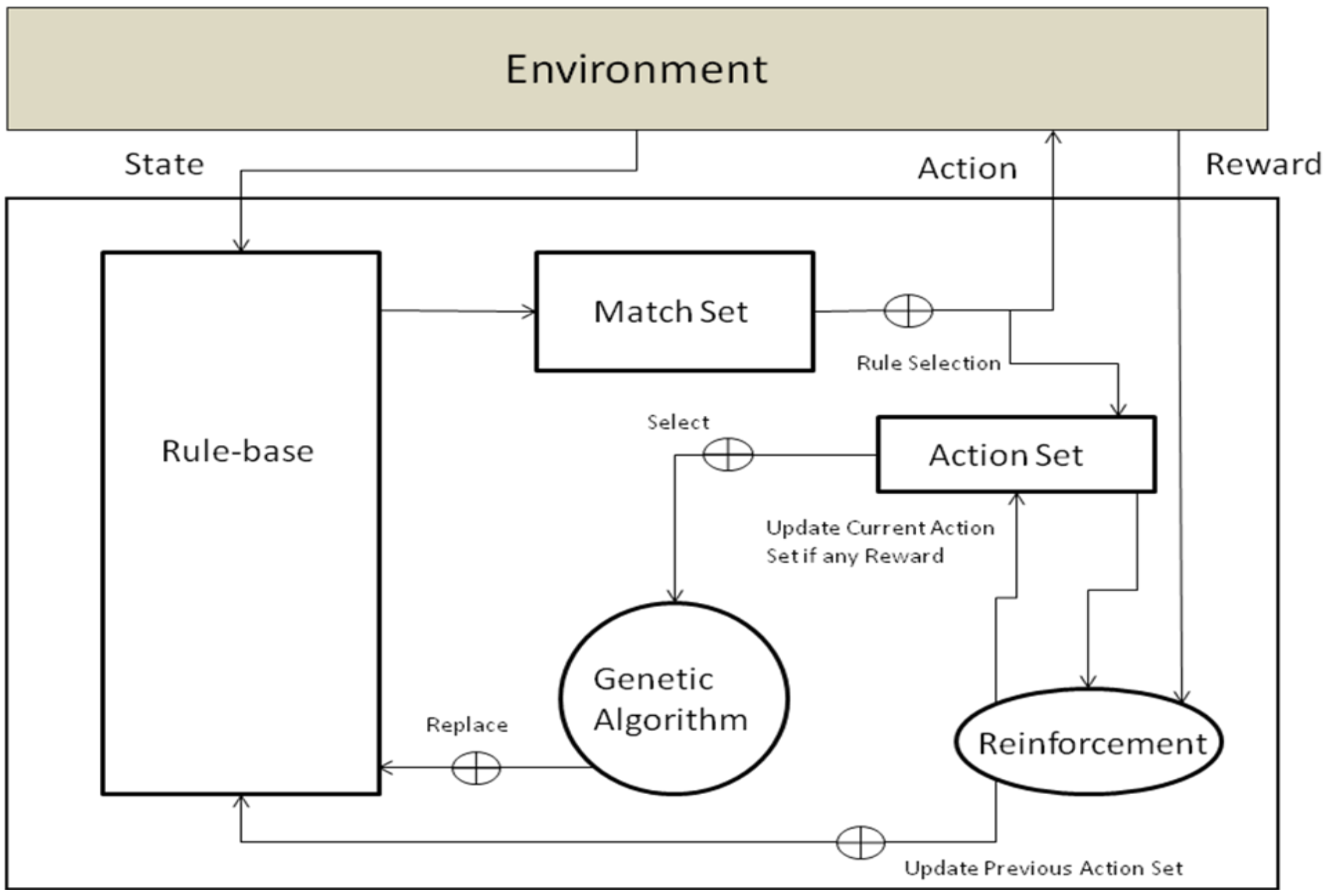

**Figure 5**: Schematic of Wilson's XCS.

In short, in XCS fitness is an inverse function of the error in reward prediction, with errors below $\varepsilon_0$ not reducing fitness. The maximum $P(a_i)$ of the system's prediction array is discounted by a factor $\gamma$ and used to update rules from the previous time step. Thus XCS exploits a form of Q-learning [Watkins, 1989] in its reinforcement procedure. The GA originally occurred in [M] but Wilson [1998] later move it to [A] to further reduce the potential for recombining rules inappropriately, i.e., when there is significant asymmetry in the generalisation space for each action in a given match set (see [Bull, 2014] for discussion). Two rules are selected based on fitness from within the chosen [A]. Rule replacement is global and based on the estimated size of each action set a rule participates in with the aim of balancing resources across niches. The GA is triggered within a given action set based on the average time since the members of the niche last participated in a GA (after [Booker, 1989]). See [Butz & Wilson, 2002] for a full algorithmic description of XCS.

Wilson originally demonstrated results on multiplexer functions and a maze problem. Importantly, he shows how maximally general solutions are evolved by XCS. This is explained by his "generalization hypothesis":

“Consider two classifiers C1 and C2 having the same action, where C2’s condition is a generalization of C1’s. …. Suppose C1 and C2 are equally accurate in that their values of $\varepsilon$ are the same. Whenever C1 and C2 occur in the same action set, their fitness values will be updated by the same amounts. However, since C2 is a generalization of C1, it will tend to occur in more [niches] than C1. Since the GA occurs in [niches], C2 would have more reproductive opportunities and thus its number of exemplars would tend to grow with respect to C1’s. …. C2 would displace C1 from the population” [Wilson, 1995].

Butz et al. [2004] studied this hypothesis formally, introducing the concept of different pressures acting within XCS and then examined how they interact. They term the process described by Wilson above as the set pressure, which occurs due to the niche GA for reproduction and global GA for deletion. Kovacs [1997] extended Wilson’s idea, presenting the “optimality hypothesis” which suggests that due to the set pressure, XCS has the potential to evolve a complete, accurate and maximally general (compact) description of a state-action-reward space. Butz et al. [2004] begin by approximating the average specificity of an action set s([A]) given the average specificity in the population s([P]):

$$s([A]) = s([P]) / (2 - s([P]))$$

For an initially random population, this indicates that the average specificity of a given [A] is lower than that of the population [P]. Opposing the set pressure are the pressures due to fitness and mutation since the former represses the reproduction of inaccurate overgeneral rules and the latter increases specificity. They then extend the set pressure definition to include the action of mutation, resulting in the "specificity equation":

$$s([P(t+1)]) = s([P(t)]) + f_{ga} \; ( ( 2 \,.\, ( s([A]) + \delta_{mut} - s([P(t)]) ) / N )$$

where $\delta_{mut}$ is the average change in specificity between a parent rule (*cl*) of specificity *s(cl)* and its offspring under mutation, defined as $0.5\mu(2 - 3s(cl))$, and $f_{ga}$ is the frequency of GA application per cycle. It is shown that, for a number of simple scenarios such as a random Boolean function, this equation is a good predictor of resulting specificity and they note this "represent[s] the first theoretical confirmation of Wilson’s generalization hypothesis" [ibid.]. The ability of XCS to maintain niches was explored formally in [Butz et al., 2007].

Butz et al. [2004] also identified two potentially conflicting challenges for XCS, namely that the initial population of rules needs to be sufficiently general to cover the input space, whilst rules must be specific enough such that there is an effective fitness gradient towards accuracy. Butz [2006] later showed how, by giving consideration to the bounds of these challenges, together with those of reproduction and niche support, XCS can PAC-learn a sub-class of *k-DNF* problems, i.e., learn their correct solution in polynomial time with high probability.

Since its introduction, a number of aspects from the wider field of machine learning have been explored within XCS (see [Lanzi, 2008] for a general review). From evolutionary computing, techniques such as rank-based selection (e.g., [Butz et al., 2005a]), parameter self-adaptation (e.g., [Hurst & Bull, 2002]), local search (e.g., [Wyatt & Bull, 2004]), and estimation of distribution algorithms (e.g., [Butz et al., 2006]) have been explored, along with non-binary representation schemes. The conditions in XCS have been represented by things such as real-valued intervals (e.g., see [Stone & Bull, 2003] for discussions), trees (e.g., [Lanzi & Perrucci, 1999], after a proposal in [Wilson, 1995]) and developmental approaches [Wilson, 2008]. The actions of rules have been represented by trees (e.g., [Iqbal et al., 2013], after [Ahluwalia & Bull, 1999]) and linear approximators (e.g., [Tran et al., 2007], after [Wilson, 2002]). A whole rule in XCS has also been represented using fuzzy logic (e.g., [Casillas et al., 2007], after [Velenzuela-Rendon, 1991]), neural networks (e.g., [Bull & O'Hara, 2002]), and logic networks (e.g., [Bull, 2009]). Techniques considered from reinforcement learning include gradient descent [Butz et al., 2005b] and eligibility traces [Drugowitsch & Barry, 2005]. Moreover, general ideas such as the use of ensembles (e.g., [Bull, et al. 2007]) and multi-agent systems (e.g., [Hercog & Fogarty, 2002]) have also been considered with XCS.

Wilson removed the message list from LCS in his ANIMAT and didn't return to the concept until he presented ZCS. As a possible area of future research, Wilson [1994] describes a "memory register". Here a global internal register's current content/state would be matched by a defined part of each rule's condition, along with the external stimulus. Similarly, rule actions would contain an element to update the content of the register, as well as supply the external response. As such, this is very similar to the original structure of CS-1. Lanzi and Wilson [2000] showed it was possible to solve non-Markov mazes through the development of the idea in XCS. An alternative approach to memory was suggested by Wilson and Goldberg [1989] wherein rules link together to form "corporations". Tomlinson and Bull (e.g., [2002]) showed some success with the idea in XCS.

Wilson prophetically suggests at the end of his paper introducing XCS: "The results point to the conclusion that accuracy-based fitness and a niche GA form a promising foundation for future classifier system research" [Wilson, 1995]. As mentioned above and shown in Figure 1, XCS and these key features have been extended from reinforcement learning as will now be discussed.

## 4. UCS: Supervised Learning

Starting with BOOLE, it has long been noted that in the use of LCS for tasks where there is an immediate reward indicating correctness, the standard reinforcement learning approach can be altered. UCS [Bernado Mansilla & Garrell, 2003] uses the accuracy calculation of AU-BOOLE to replace the standard running average error update in XCS. That is, $\kappa_j$ = number of correct classifications/experience. Where experience is the number of times a rule has matched. Thereafter, $F_j = (\kappa_j)^{v}$ and the GA is run in the correct set [C], with deletion a global operation based upon the size of [C] (Figure 6).

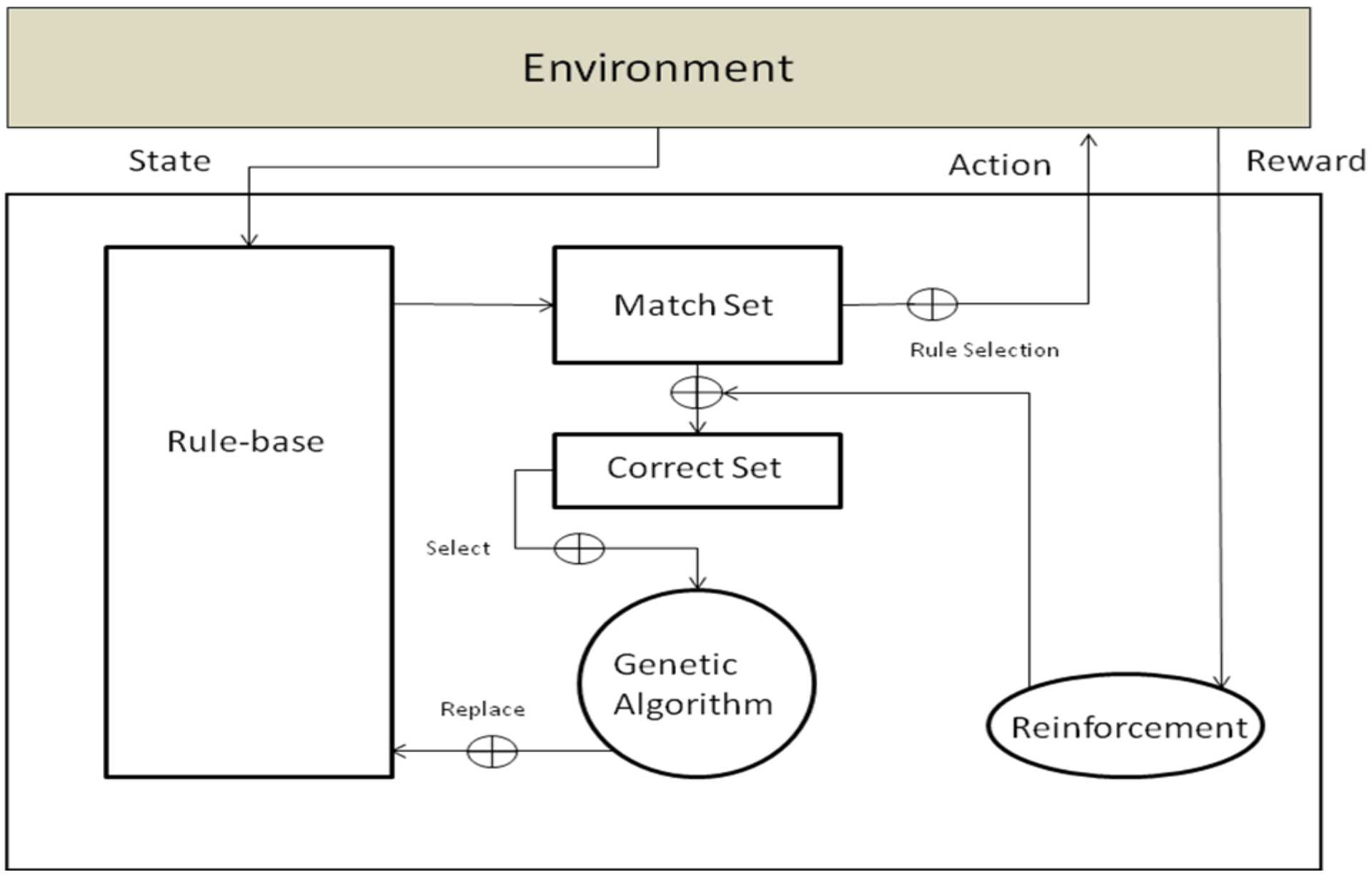

**Figure 6**: Schematic of UCS.

The main effect of the change to a supervised update is that UCS only maintains a set of rules which receive high payoff, as opposed to XCS's construction of a full state-action-reward map. As a consequence, UCS was shown to learn more quickly than an equivalent XCS on a number of benchmark tasks. As well as the reduced generalization task, it was also shown to learn more effectively due to the change in the fitness pressure for certain types of problem from the simplified fitness function. UCS and XCS are shown to be competitive with a number of well-known machine learning techniques over well-known real-world datasets.

In a few cases, XCS was found to outperform UCS on the real-world datasets and it was speculated this is due to in part to a lack of fitness sharing within niches. Later inclusion of the same relative accuracy calculation into UCS gives improved performance, particularly with unbalanced datasets [Orriols-Puig & Bernado Mansilla, 2008]. However, XCS remains a robust classification data mining algorithm (see [Fernandez et al., 2010]).

Like XCS, a number of techniques have been incorporated into UCS, such as the use of rank-based selection (e.g., [Orriols-Puig & Bernado Mansilla, 2008]) and fuzzy logic (e.g., [Orriols-Puig et al., 2009]). It has also been used within ensembles, including with the use of neural networks to provide the action (e.g., [Dam et al., 2008]). Ideas from the wider ensemble/mixture-of-experts literature have also been used to understand and refine UCS (e.g., [Edakunni et al., 2011]).

## 5. XCSC: Unsupervised Learning

Unsupervised learning describes those tasks under which structure is sought in unlabelled data without further external input. Perhaps somewhat surprisingly, no previous suggestion of the use of LCS for such learning is known in the literature until the work of Tammee et al. [2006; 2007] on clustering (see Figure 7). Clustering is an important unsupervised learning technique where a set of data are grouped into clusters in such a way that data in the same cluster are similar in some sense and data in different clusters are dissimilar in the same sense (see [Xu & Wunsch, 2009] for an overview). Most clustering algorithms require the user to provide the number of clusters, and the user in general has no idea about the number of clusters (e.g., see [Tibshirani et al., 2000]). Hence this typically results in the need to make several clustering trials with different numbers of clusters from 1 to the square-root of the number of data points, and select the best clustering among the partitioning with different number of clusters.

Tammee et al. show how the generalization mechanisms of XCS can be used to identify clusters – both their number and description. Rules in their XCSC use an interval representation of the form $\{\{c_1, s_1\}, \ldots\ \{c_d, s_d\}\}$, where $c$ is the interval's range centre from [0.0,1.0] and $s$ is the "spread" from that centre from the range (0.0,$s_0$] and $d$ is a number of dimensions. Each interval predicates' upper and lower bounds are calculated as follows: [$c_i$ - $s_i$, $c_i$ + $s_i$]. If an interval predicate goes outside the problem space bounds, it is truncated. Rule fitness consists of updating the matching error ε which is derived from the Euclidean distance with respect to the input $x$ and $c$ in the condition of each member of the current [M] using the Widrow-Hoff delta rule with learning rate β:

$$\varepsilon_j \leftarrow \varepsilon_j + \beta\left( \left(\left(\sum_{l=1}^{d}(x_l - c_{lj})^2\right)\right)^{1/2} - \varepsilon_j \right)$$

The rest of XCS processing remains unchanged. Hence the set pressure encourages the evolution of rules which cover many data points and the fitness pressure acts as a limit upon the separation of such data points, i.e., the error.

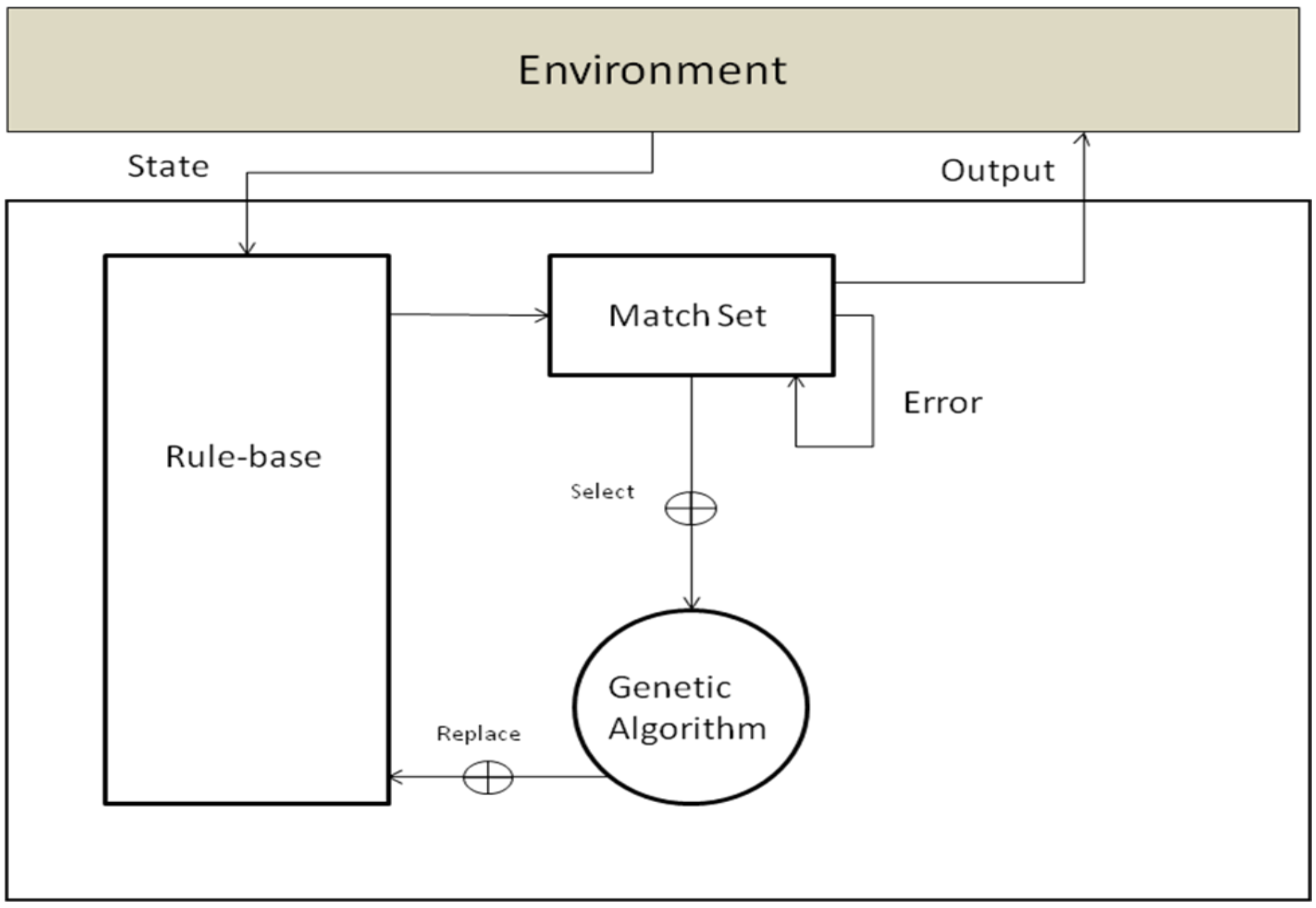

**Figure 7**: Schematic of XCSC.

Tammee et al. [2006] began by using a slightly simplified version of XCS as the underlying LCS (YCS) [Bull, 2005], but found that XCS's relative accuracy fitness function was more effective than a function directly inversely proportional to error [Tammee et al., 2007]. Note this is similar to the aforementioned findings with UCS [Orriols-Puig & Bernado Mansilla, 2008]. Moreover, since the $\varepsilon_0$ parameter controls the error threshold of rules, Tammee et al. investigated the sensitivity of XCSC to its value by varying it. Their experiments show that, if $\varepsilon_0$ is set high, e.g., 0.1, in less-separated data the contiguous clusters are covered by the same rules. They therefore developed an adaptive threshold parameter scheme which uses the average error of the current [M]:

$$\varepsilon_0 = \tau(\sum \varepsilon_j / N_{[M]})$$

Where $\varepsilon_j$ is the average error of each rule in the current match set and $N_{[M]}$ is the number of rules in the current match set. This is applied before the fitness function calculations. Experimentally Tammee et al. found $\tau$=1.2 was most effective for the problems they considered.

This work has recently been extended to include hierarchical cluster/rule merging and voting (e.g., [Qian et al., 2013]).

## 6. XCSF: Function Learning

Wilson [1995] proposed that XCS could be modified to learn functions, i.e., problems of the general form $y=f(x)$, and subsequently presented XCSF [Wilson, 2002]. Rules in XCSF typically use an interval representation of the form $\{\{l_1\ u_1\}, \ldots\ \{l_d, u_d\}\}$, where $l_i$ ("lower") and $u_i$ ("upper") are integers. A rule matches an input $x$ with attributes $x_i$ if and only if $l_i \le x_i \le u_i$ for all $x_i$. After first using the standard prediction creation of XCS, Wilson introduces piecewise-linear approximators to each rule, i.e., functions of the form $h(x) = w_0 + w_1x_1 + \ldots w_dx_d$. Rather than add the weights $w_j$ to the rule representation to be learned under the GA, a variant of a simple gradient descent method is employed: $\Delta w_j = (\eta/|x'|^2)\ (t - o)x_j$, where $t$ is the target, $o$ is the output, and $\eta$ is a learning rate.

Wilson shows the basic XCSF learning a sine function and multi-dimension root-mean-squared functions. This has subsequently been explored extensively, using a variety of rule condition representations and function approximation techniques (e.g., see [Lanzi et al., 2007][Butz et al., 2008]). The theoretical underpinnings of XCS have also been extended to XCSF (e.g., [Stalph et al., 2012a]).

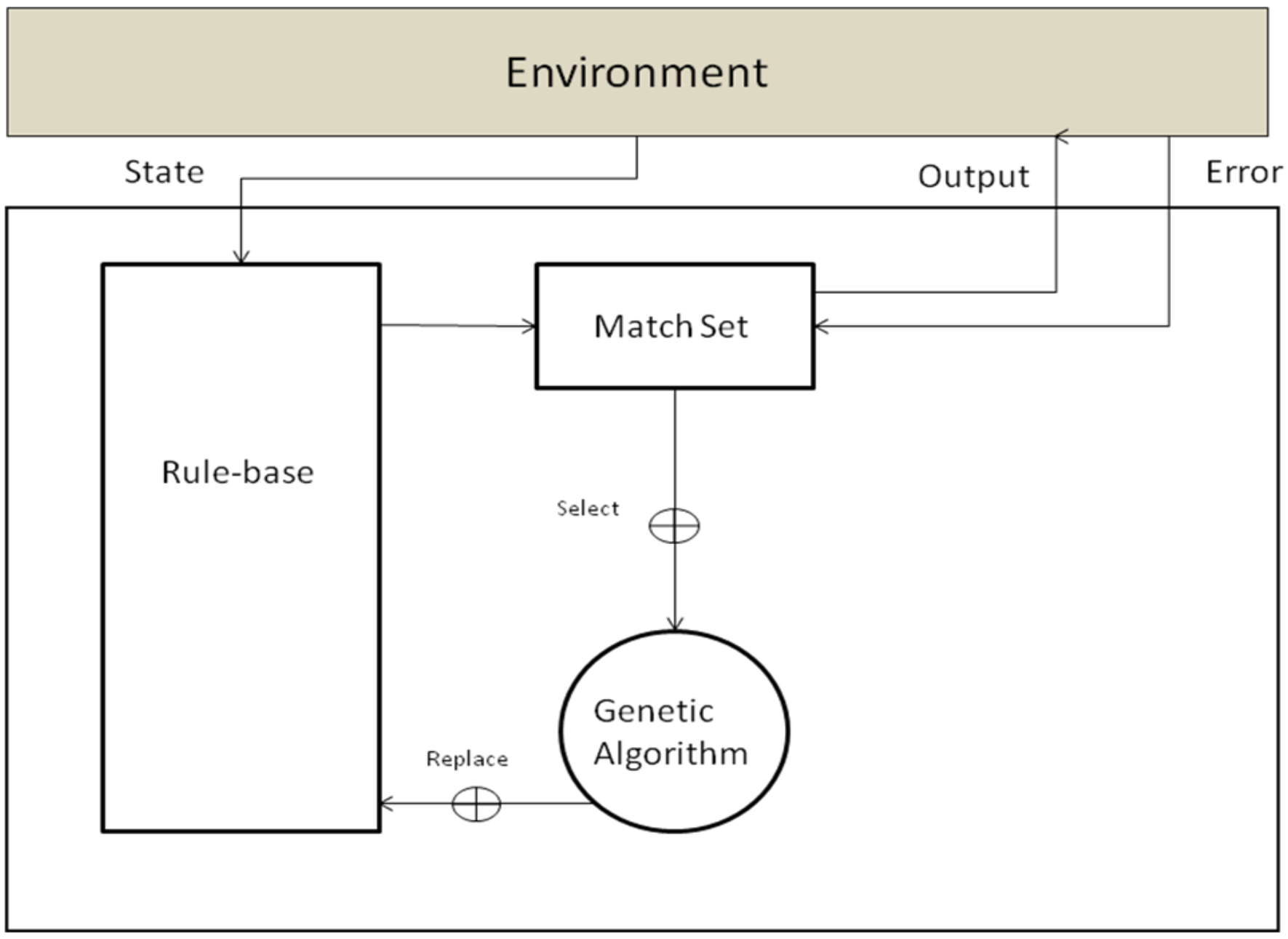

**Figure 8**: Schematic of XCSF.

Wilson [2002] extended the idea to propose a generalized rule format such that the prediction is computed in the same way, as opposed to maintained as (an adjusted) parameter. Again, this has been explored using a variety of rule representations and approximators (e.g., see [Loiacono et al., 2009]). It can be noted that in rule representations which can also provide memory through individual rule-internal structures, such as recurrent connections in a network, this opens up new ways by which to solve non-Markov tasks (e.g., see [Preen & Bull, 2013]).

## 7. XACS: Model Learning

As noted above, Stolzmann [1998] presented an accuracy-based LCS in which rules are extended to predict the subsequent sensory state from their use. That is, rules are of the general form "IF condition AND assertion THEN effect". The mechanism through which such rules are learned is based upon the theory of Anticipatory Behavioural Control [Hoffmann, 1993] and not a simulated evolutionary process. The search algorithm, termed the Anticipatory Learning Process (ALP), has thus far relied upon the traditional ternary alphabet {0,1,#}. Butz et al. introduced the use of a niched GA alongside the ALP to improve the generalization abilities of ACS, termed ACS2 (see [Butz, 2002]). Whilst effective, ACS2 was found to sometimes struggle to form both accurate environment models and state-action-reward models simultaneously. Drawing on XCSF, ACS2 was subsequently extended in XACS [Butz & Goldberg, 2003] (figure 9).

XACS maintains the principle features of ACS(2), using the ALP to specialize rules when their anticipated effect does not match the next state. As in ACS2, a # symbol in an effect indicates that the bit is not anticipated to change in the next state, whereas defined bits are anticipated to change to that value (unlike ACS). The GA is the same as in ACS2, using the time triggered scheme of XCS with the mutation process only introducing #'s and crossover only happening over conditions. Running alongside ACS2 is a variant of XCSF to learn the value of states. The rules consist of a condition and prediction parameter, as in the first version of XCSF described above. The XCSF component is used each time external reward is received from the environment, updating predicted values using both its own current prediction and the model knowledge of the ACS2 component.

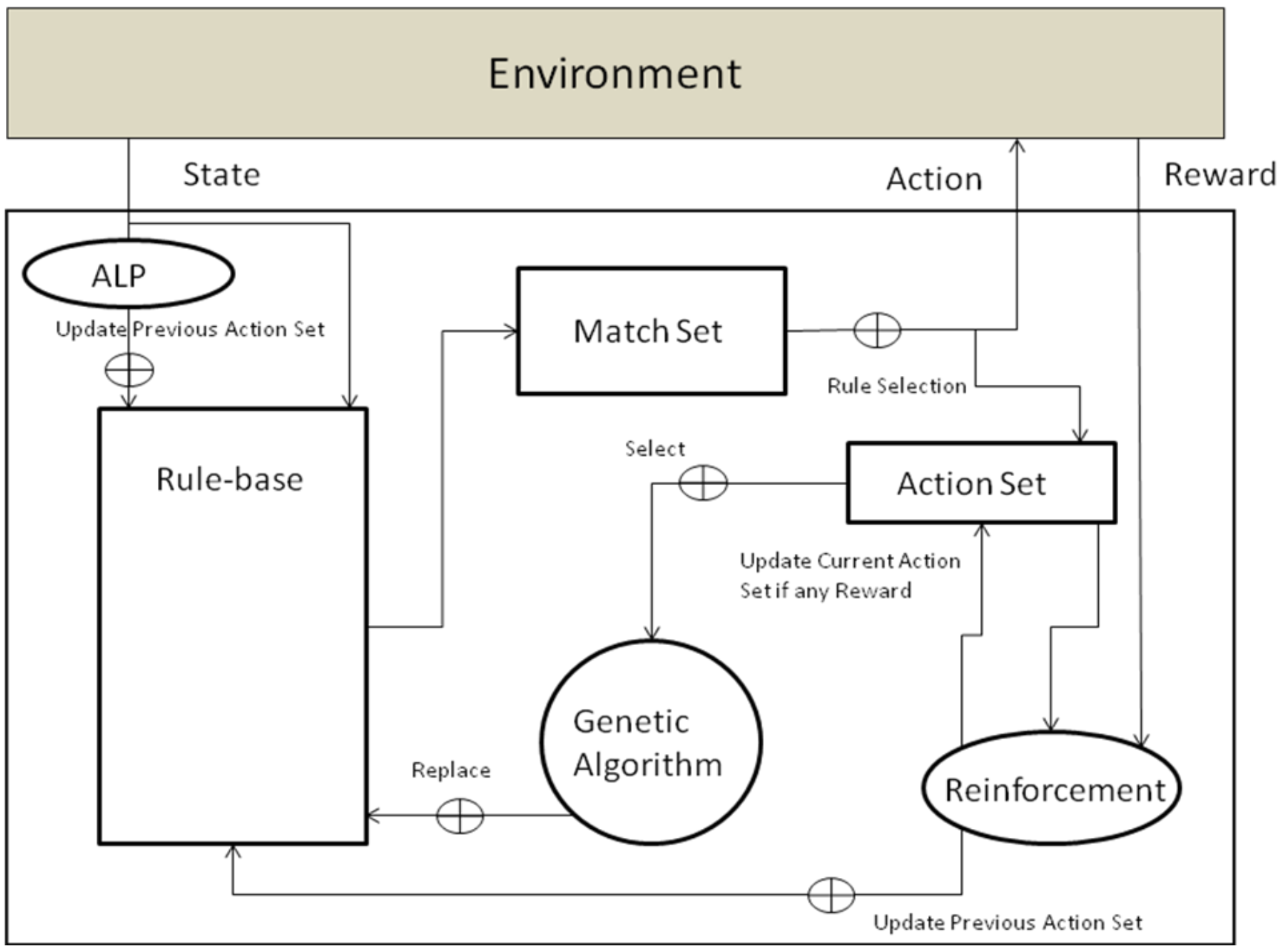

**Figure 9**: Schematic of XACS (without XCSF component).

Butz and Goldberg [2003] report improved performance over ACS2 in blocks world problems of varying sizes. Benefits of the model include a mechanism through which to bias action selection such that those rules whose anticipations are least accurate are chosen preferentially over a random action, the ability to learn multiple tasks simultaneously over the same environment by including an XCSF component per task (see [Studley & Bull, 2006] for a related study), etc. Given the supervised learning-like nature of building anticipations, they have also been learned in a version of XCSF using a neural network to predict the next state (e.g., [Bull et al., 2007]).

## 8. Conclusion

Architecturally, XCS can be traced from ANIMAT via ZCS, with GOFER's triggered niche GA being included. The use of accuracy began with CS-1, although it was focused on the highest reward per niche. Moreover, XCS's generalization pressure shares features with that in CS-1 since it is also based on accuracy and rate of use. In CS-1, predicted rewards are only updated if they are accurate or below the current estimate, with action and GA selection based upon this parameter: more accurate rules are more likely to reproduce. Rule ages are reset after use and deletion is based upon age: more frequently used rules are less likely to be replaced. Thus accurate, more general (frequently used) rules are propagated in CS-1. XCS combines both accuracy, in its pure form, and frequency of use into the selection process of the GA. This creates a generalization pressure but, importantly, also frees the deletion process of the GA to be used to maintain multiple niches in an emergent way thereby addressing one of the main issues in CS-1 that Holland sought to tackle by switching to strength sharing in his subsequent LCS. Much has subsequently been explored with XCS, and there remains much to explore.

XCS has been used effectively to control physical robots in continuous time and space where the action space was discrete and relatively small (e.g., [Studley & Bull, 2005]). Wilson [2007] has presented a "generalized classifier system" concept whereby LCS can work in a continuous-valued action space. Whilst studies have shown progress in this area for regression problems (e.g., [Preen & Bull, 2013]), there is still much to be done for reinforcement learning problems (e.g., [Casillas et a., 2007][Howard et al., 2009]).

As highlighted in [Bull, 2011], XCSC can in hindsight be viewed as a type of Artificial Immune System (AIS). For nearly thirty years (starting with [Farmer et al., 1986]) similarities between LCS and AIS have periodically been noted, but the two fields have developed independently. The use of selection within niches of co-active rules is akin to the scheme used in a general class of AIS known as clonal selection algorithms (e.g., CLONALG [De Castro & Von Zuben, 2002]). The use of a time-delayed evolutionary process is also similar to the dendritic cell AIS (e.g., [Greensmith et al., 2008]). It can also be noted that the adaptive affinity threshold finding in XCSC [Tammee et al., 2007] is much like the result reported in [Bezerra et al., 2005] with an AIS. A potential area for future research would therefore appear to be to explore the cross-fertilization of mechanisms between what are now two relatively mature fields (e.g., see [Timmis et al., 2008] for an overview of AIS).

LCS were presented as an architecture through which to study cognitive systems. Whilst the reinforcement learning element has a clear connection to neuroscience (e.g., [Schultz, 1998]), the use of an evolutionary process to build knowledge representations has lacked a strong connection. The general similarities between LCS and artificial neural networks have long been noted (e.g., [Smith & Cribbs, 1994]), and as mentioned above such networks have been used as rules, including spiking models (e.g. [Howard et al., 2010]). However, there are suggestions that neurogenesis may be significant in adult learning (e.g., [Becker, 2005]) and that such neurons may vary genetically upon production (e.g., [Coufal et al., 2009]). Selectionist models of brains, i.e., forms of neural Darwinism, continue to be developed (e.g., [Fernando et al., 2012]). It has recently been shown [Stalph et al., 2012b] that XCSF can be very similar to the locally-weighted projection regression algorithm [Vijayakumar et al., 2005], suggesting its rules may be seen to specify local receptive fields. Another potential area for future research would therefore appear to be to move LCS closer to computational neuroscience.

## References

Ahluwalia, M. & Bull, L. (1999) A Genetic Programming-based Classifier System. In W. Banzhaf et al. (eds) *GECCO-99: Proceedings of the Genetic and Evolutionary Computation Conference.* Morgan Kaufmann, pp11-18.

Becker, S. (2005) A Computational Principle for Hippocampal Learning and Neurogenesis. *Hippocampus* 15 (6): 722–38.

Bernado Mansilla, E. & Garrell, J. (2003) Accuracy-Based Learning Classifier Systems: Models, Analysis and Applications to Classification Tasks. *Evolutionary Computation* 11(3): 209-238.

Bezerra, G., Barra, T., de Castro, L. & Von Zuben (2005) Adaptive Radius Immune Algorithm for Data Clustering. In C. Pilat et al. (eds) *Proceedings of the 4th International Conference on Artificial Immune Systems*. Springer, pp.290-303.

Bonelli, P., Parodi, A., Sen, S. & Wilson, S.W. (1990) NEWBOOLE: A Fast GBML System. In *International Conference on Machine Learning*. Morgan Kaufmann, pp153-159.

Booker, L. (1982) *Intelligent Behavior as an Adaptation to the Task Environment.* Ph.D. Thesis, the University of Michigan.

Booker, L.B. (1985) Improving the Performance of Genetic Algorithms in Classifier Systems. In J.J. Grefenstette (ed) *Proceedings of the First International Conference on Genetic Algorithms and their Applications.* Lawrence Erlbaum Associates, pp80-92.

Booker, L. (1988) Classifier Systems that Learn Internal World Models. *Machine Learning* 3: 161-192.

Booker, L. (1989) Triggered Rule Discovery in Classifier Systems. In J. Schaffer (ed) *Proceedings of the International Conference on Genetic Algorithms*. Morgan Kaufmann, pp265-274.

Box, G. (1957) Evolutionary Operation: A Method for Increasing Industrial Productivity. *Journal of Royal Statistical Society C* 6(2): 81-101.

Bull, L. (2004)(ed) *Applications of Learning Classifier Systems*. Springer.

Bull, L. (2005) Two Simple Learning Classifier Systems. In L. Bull & T. Kovacs (eds) *Foundations of Learning Classifier Systems*. Springer, pp63-90.

Bull, L. (2009) On Dynamical Genetic Programming: Simple Boolean Networks in Learning Classifier Systems. *International Journal of Parallel, Emergent and Distributed Systems* 24(5): 421-442

Bull, L. (2011) Towards a Mapping of Modern AIS and LCS. In P. Lio et al. (eds) *Proceedings of the Tenth International Conference on Artificial Immune Systems*. Springer, pp371-382

Bull, L. (2014) Exploiting Generalisation Symmetries in Accuracy-based Learning Classifier Systems: An Initial Study. http://arxiv.org/abs/1401.2949

Bull, L. & Hurst, J. (2002) ZCS Redux. *Evolutionary Computation* 10(2): 185-205.

Bull, L. & O'Hara, T. (2002) Accuracy-based Neuro and Neuro-Fuzzy Classifier Systems. In W.B.Langdon et al. (eds) *GECCO-2002: Proceedings of the Genetic and Evolutionary Computation Conference.* Morgan Kaufmann, pp905-911

Bull, L. & Kovacs, T. (2005)(eds) *Foundations of Learning Classifier Systems*. Springer.

Bull, L., Lanzi, P-L. & O'Hara, T. (2007) Anticipation Mappings for Learning Classifier Systems. In *Proceedings of the IEEE Congress on Evolutionary Computation*. IEEE Press, pp2133-2140

Bull, L., Studley, M., Bagnall, A. & Whittley, I. (2007) Learning Classifier System Ensembles with Rule Sharing. *IEEE Transactions on Evolutionary Computation* 11(4): 496-502

Bull, L., Bernado Mansilla, E. & Holmes, J. (2008)(eds) *Learning Classifier Systems in Data Mining*. Springer.

Butz, M. (2002) *Anticipatory Learning Classifier Systems*. Kluwer.

Butz, M.V. (2006) *Rule-based Evolutionary Online Learning Systems*. Springer.

Butz, M.V. & Wilson, S.W. (2002) An algorithmic description of XCS. *Soft Computing 6*(3-4): 144-153

Butz, M.V. & Goldberg, D.E. (2003) Generalized State Values in an Anticipatory Learning Classifier System. In Butz, M.V., Sigaud, O., & G´erard, P. (eds) *Anticipatory Behavior in Adaptive Learning Systems*. Springer, pp. 282–301.

Butz, M.V., Kovacs, T., Lanzi, P-L & Wilson, S.W. (2004) Toward a Theory of Generalization and Learning in XCS. *IEEE Transactions on Evolutionary Computation* 8(1): 28-46

Butz, M.V., Sastry, K. & Goldberg, D.E. (2005a) Strong, Stable, and Reliable Fitness Pressure in XCS due to Tournament Selection. *Genetic Programming and Evolvable Machines* 6(1): 53-77

Butz, M.V., Goldberg, D. E. & Lanzi, P-L (2005b) Gradient Descent Methods in Learning Classifier Systems: Improving XCS Performance in Multi-step Problems. *IEEE Transactions on Evolutionary Computation* 9(5): 452-473

Butz, M.V., Pelikan, M., Llora, X. & Goldberg, D.E. (2006) Automated Global Structure Extraction for Effective Local Building Block Processing in XCS. *Evolutionary Computation* 14(3): 345-380

Butz, M.V., Goldberg, D., Lanzi, P-L. & Sastry, K. (2007) Problem Solution Sustenance in XCS: Markov Chain Analysis of Niche Support Distributions and the Impact on Computational Complexity. *Genetic Programming and Evolvable Machines* 8(1): 5-37

Butz, M.V., Lanzi, P-L. & Wilson, S.W. (2008) Function Approximation with XCS: Hyperellipsoidal Conditions, Recursive Least Squares, and Compaction. *IEEE Trans. Evolutionary Computation 12*(3): 355-376.

Casillas, J. Carse, B. & Bull, L. (2007) Fuzzy XCS: a Michigan Genetic Fuzzy System. *IEEE Transactions on Fuzzy Systems* 15(4): 536-550

Cliff, D. & Ross, S. (1995) Adding Temporary Memory to ZCS. *Adaptive Behavior* 3(2): 101-150.

Coufal, N. et al (2009) L1 Retrotransposition in Human Neural Progenitor Cells. *Nature* 460: 1127-1131.

Dam, H., Abbass, H., Lokan, C. & Yao, X. (2008) Neural-Based Learning Classifier Systems. *IEEE Transactions on Knowledge Data Engineering* 20(1): 26-39

De Castro, L. & Von Zuben, F. (2002) Learning and Optimization using the Clonal Selection Principle. *IEEE Transactions on Evolutionary Computation* 6(3): 239-251.

Dorigo, M. & Bersini, H. (1994) A Comparison of Q-learning and Classifier Systems. In D. Cliff, P. Husbands, J-A. Meyer & S. W. Wilson (eds) *From Animals to Animats 3: Proceedings of the Third International Conference on Simulation of Adaptive Behaviour.* MIT Press, pp248-255.

Drugowitsch, J. & Barry, A. (2005) XCS with Eligibility Traces. In H.G. Beyer et al. (eds) *GECCO-2005: Proceedings of the Genetic and Evolutionary Computation Conference.* Morgan Kaufmann, pp1851-1858

Edakunni, N., Brown, G. & Kovacs, T. (2011) Online, GA based Mixture of Experts : a Probabilistic Model of UCS. In *GECCO-2011: Proceedings of the Genetic and Evolutionary Computation Conference*. ACM Press, pp1267–1274.

Eiben, A. & Smith, J. (2003) *Introduction to Evolutionary Computing*. Springer.

Farley, B. & Clark, W. (1954) Simulation of Self-organizing Systems by Digital Computer. *IRE Transactions on Information Theory* 4: 76-84.

Farmer, J.D., Packard, N., & Perelson, A. (1986) The Immune System, Adaptation and Machine Learning. *Physica D* 22: 187-204.

Fernández, A., García, S., Luengo, J., Bernadó-Mansilla, E., & Herrera, F. (2010) Genetics-Based Machine Learning for Rule Induction: State of the Art, Taxonomy, and Comparative Study. *IEEE Transactions on Evolutionary Computation* 14(6): 913-941

Fernando, C., Szathmary, E. & Husbands, P. (2012) Selectionist and Evolutionary Approaches to Brain Function: A Critical Appraisal. *Frontiers in Computational Neuroscience* 6(24).

Fraser, A. (1957) Simulation of Genetic Systems by Automatic Digital Computers. I. Introduction. *Australian Journal of Biological Sciences* 10: 484-491.

Frey, P. & Slate, D. (1991) Letter Recognition using Holland-style Adaptive Classifiers. *Machine Learning* 6: 161-182.

Goldberg, D. (1985) Genetic Algorithms and Rule Learning in Dynamic System Control. In J.J. Grefenstette (ed) *Proceedings of the First International Conference on Genetic Algorithms and their Applications.* Lawrence Erlbaum Associates, pp8-15.

Greensmith, J., Feyereisl, J. & Aickelin, U. (2008) DCA: Some Comparison. *Evolutionary Intelligence* 1(2): 85-112.

Hartley, A. (1999) Accuracy-based Fitness Allows Similar Performance to Humans in Static and Dynamic Classification Environments. In W. Banzhaf et al. (eds) *GECCO-99: Proceedings of the Genetic and Evolutionary Computation Conference*. Morgan Kaufmann, pp266-273.

Hercog, L. & Fogarty, T.C. (2002) Coevolutionary Classifier Systems for Multi-Agent Simulation. In *Proceedings of the IEEE Congress on Evolutionary Computation*. IEEE Press, pp1798-1803.

Holland, J.H. (1975) *Adaptation in Natural and Artificial Systems*. University of Michigan Press.

Holland, J.H. (1976) Adaptation. In Rosen & Snell (eds) *Progress in Theoretical Biology*, 4. Plenum, pp263-293.

Holland, J.H. (1980) Adaptive Algorithms for Discovering and using General Patterns in Growing Knowledge Bases. *International Journal of Policy Analysis and Information Systems* 4(3): 245-268.

Holland, J. H. (1985) Properties of the Bucket Brigade. In J.J. Grefenstette (ed) *Proceedings of the First International Conference on Genetic Algorithms and their Applications*. Lawrence Erlbaum Associates, pp1-7.

Holland, J.H. (1986). Escaping brittleness: the possibilities of general-purpose learning algorithms applied to parallel rule-based systems. In Michalski, Carbonell, & Mitchell (eds) *Machine learning, an artificial intelligence approach*. Morgan Kaufmann, pp593-623.

Holland, J.H. (1990) Concerning the Emergence of Tag-Mediated Lookahead in Classifier Systems. *Physica D* 42: 188-201.

Holland, J.H. & Reitman, J.H. (1978) Cognitive Systems Based in Adaptive Algorithms. In Waterman & Hayes-Roth (eds*) Pattern-directed Inference Systems*. Academic Press, pp313-329.

Holland, J.H., Holyoak, K.J., Nisbett, R.E. & Thagard, P.R. (1986) *Induction: Processes of Inference, Learning and Discovery*. MIT Press.

Howard, D., Bull, L. & Lanzi, P-L. (2009) Continuous Actions in Continuous Space and Time using Self-Adaptive Constructivism in Neural XCSF. In *GECCO-2009: Proceedings of the Genetic and Evolutionary Computation Conference.* ACM Press, pp1219-1226.

Howard, D., Bull, L. & Lanzi, P-L. (2010) A Spiking Neural Representation for XCSF. In *Proceedings of the IEEE Congress on Evolutionary Computation.* IEEE Press, pp1-8.

Hurst, J. & Bull, L. (2002) A Self-Adaptive XCS. In P-L. Lanzi, W. Stolzmann & S.W. Wilson (eds) *Advances in Learning Classifier Systems: Proceedings of the Fourth International Workshop on Learning Classifier Systems*. Springer, pp57-73.

Iqbal, M., Browne, W. & Zhang, M. (2013) Evolving optimum populations with XCS classifier systems - XCS with code fragmented action. *Soft Computing* 17(3): 503-518

Iqbal, M., Browne, W. & Zhang, M. (2014) Reusing Building Blocks of Extracted Knowledge to Solve Complex, Large-Scale Boolean Problems. *IEEE Transactions on Evolutionary Computation*.

Kovacs, T. (1997) XCS Classifier System Reliably Evolves Accurate, Complete, and Minimal Representations for Boolean Functions. In Roy, Chawdhry & Pant (eds) *Soft Computing in Engineering Design and Manufacturing*. Springer, pp59–68.

Lanzi, P-L. (2008) Learning Classifier Systems: Then and Now. *Evolutionary Intelligence* 1(1): 63-82.

Lanzi, P-L & Perrucci, A. (1999) Extending the Representation of Classifier Conditions Part II: From Messy Coding to S-Expressions. In W. Banzhaf et al. (eds) *GECCO-99: Proceedings of the Genetic and Evolutionary Computation Conference*. Morgan Kaufmann, pp345-352.

Lanzi, P-L & Riolo, R. (2000) A Roadmap to the Last Decade of Learning Classifier System Research. In P-L. Lanzi, W. Stolzmann & S.W. Wilson (eds) *Learning Classifier Systems: From Foundations to Applications*. Springer, pp33-62.

Lanzi, P-L. & Wilson, S.W. (2000) Toward Optimal Classifier System Performance in Non-Markov Environments. *Evolutionary Computation 8*(4): 393-418

Lanzi, P-L, Loiacono, D., Wilson, S.W. & Goldberg, D. (2007) Generalization in the XCSF Classifier System: Analysis, Improvement, and Extension. *Evolutionary Computation 15*(2): 133-168

Loiacono, D. & Lanzi, P-L. (2009) Recursive Least Squares and Quadratic Prediction in Continuous Multistep Problems. In J. Bacardit et al. (eds) *Learning Classifier Systems: Revised Selected Papers*. Springer, pp70-86

Orriols-Puig, A. & Bernado Mansilla, E. (2008) Revisiting UCS: Description, Fitness Sharing, and Comparison with XCS. In J. Bacardit et al. (eds) *Learning Classifier Systems: Revised Selected Papers*. Springer, pp96-116.

Orriols-Puig, A., Casillas, J. & Bernadó Mansilla, E. (2009) Fuzzy-UCS: A Michigan-Style Learning Fuzzy-Classifier System for Supervised Learning. *IEEE Transactions on Evolutionary Computation* 13(2): 260-283.

Preen, R. & Bull, L. (2013) Dynamical Genetic Programming in XCSF. *Evolutionary Computation* 21(3): 361-388.

Qian, L., Shi, Y., Gao, Y. & Yin, H. (2013) Voting-XCSc: A Consensus Clustering Method via Learning Classifier System. In H. Yin et al. (eds) *Intelligent Data Engineering and Automated Learning – IDEAL*. Springer, pp603-610.

Riolo, R. (1991) Lookahead Planning and Latent Learning in a Classifier System. In J-A. Meyer & S. W. Wilson (eds) *From Animals to Animats: Proceedings of the First International Conference on Simulation of Adaptive Behaviour.* MIT Press, pp316-326.

Roberts, G. (1993) Dynamic Planning for Classifier Systems. In S. Forrest (ed), *Proceedings of the 5th International Conference on Genetic Algorithms*. Morgan Kaufmann, pp231-237.

Samuel, A.L. (1959) Some Studies in Machine Learning using the Game of Checkers. *IBM Journal of Research and Development* 3: 211-229.

Samuel, A.L. (1967) Some Studies in Machine Learning using the Game of Checkers. II. Recent Progress. *IBM Journal of Research and Development* 11: 601-617.

Schultz, W. (1998) Predictive Reward Signal of Dopamine Neurons. *Journal of Neurophysiology* 68: 1190-1208.

Seward, J. (1949) An Experimental Analysis of Latent Learning. *Journal of Experimental Psychology* 39: 177-186.

Shannon, C. (1950) Programming a Computer for Playing Chess. *Philosophical Magazine* 41: 256-275.

Smith, S.F. (1980) *A Learning System Based on Genetic Adaptive Algorithms*. PhD Thesis, University of Pittsburgh.

Smith, R. & Cribbs, H. (1994) Is a Learning Classifier System a Type of Neural Network? *Evolutionary Computation* 2(1): 19-36

Smith, R., Jiang, M., Bacardit, J., Stout, M., Krasnogor, N. & Hirst, J. (2010) A Learning Classifier System with Mutual-Information-based Fitness. *Evolutionary Intelligence* 3(1): 31-50.

Stalph, P., Llorà, X., Goldberg, D. & Butz, M.V. (2012a) Resource Management and Scalability of the XCSF Learning Classifier System. *Theoretical Computer Science* 425: 126-141

Stalph, P., Rubinsztajin, J., Sigaud, O. & Butz, M.V. (2012b) Function Approximation with LWPR and XCSF: A Comparative Study. *Evolutionary Intelligence* 5(2): 103-116.

Stolzmann, W. (1998) Anticipatory Classifier Systems. In Koza et al. (eds.) *Genetic Programming 1998: Proceedings of the Third Annual Conference,* Morgan Kaufmann, pp658-654.

Stone, C. & Bull, L. (2003) For Real! XCS with Continuous-Valued Inputs. *Evolutionary Computation* 11(3): 299-336

Stone, C. & Bull, L. (2005) Comparing XCS and ZCS on Noisy Continuous-Valued Environments. *Technical Report: UWELCSG05-002* (http://www.cems.uwe.ac.uk/lcsg)

Studley, M. & Bull, L. (2005) X-TCS: Accuracy-based Learning Classifier System Robotics. In *Proceedings of the IEEE Congress on Evolutionary Computation*. IEEE, pp2099-2106

Studley, M. & Bull, L. (2006) Using the XCS Classifier System for Multi-objective Reinforcement Learning Problems. *Artificial Life* 13(1): 69-86

Sutton, R. & Barto, A. (1981) Toward a Modern Theory of Adaptive Networks: Expectation and Prediction. *Psychological Review* 88: 135-170.

Sutton, R. & Barto, A. (1998) *Reinforcement Learning*. MIT Press.

Tammee, K., Bull, L. & Ouen, P. (2006) A Learning Classifier System Approach to Clustering. In *Proceedings of the 6th International Conference on Intelligent Systems Design and Applications*. IEEE, pp621-626.

Tammee, K., Bull, L. & Ouen, P. (2007) Towards Clustering with XCS. In D. Thierens et al. (eds) *GECOO-2007: Proceedings of the Genetic and Evolutionary Computation Conference.* ACM Press, pp1854-1860.

Thorndike, E. (1911) *Animal Intelligence.* Macmillan Company.

Tibshirani, R., Walther, G., & Hastie, T. (2000) Estimating the Number of Clusters in a Dataset Via the Gap Statistic. *Journal of the Royal Statistical Society*, B, 63: 411-423.

Timmis, J., Andrews, P., Owens, N. & Clark, E. (2008) An Interdisciplinary Perspective on Artificial Immune Systems. *Evolutionary Intelligence* 1(1): 5-26

Tomlinson, A. & Bull, L. (2002) An Accuracy-Based Corporate Classifier System. *Soft Computing* 6(3-4): 200-215

Tran, T., Sanza, C., Duthen, Y. & Nguyen, D. (2007) XCSF with Computed Continuous Action. In D. Thierens et al. (eds) *GECCO-07: Proceedings of the Genetic and Evolutionary Computation Conference*. ACM Press, pp1861-1868.

Turing, A. (1948) Intelligent Machinery. Reprinted in: Copeland, J. (2004) *The Essential Turing.* Oxford, pp395-432.

Valenzuela-Rendón, M. (1991) The Fuzzy Classifier System: A Classifier System for Continuously Varying Variables. In R. Belew & L. Booker (eds) *Proceedings of the 4th International Conference on Genetic Algorithms.* Morgan Kaufmann, pp. 346–353.

Vijayakumar, S., D'Souza, A. & Schall, S. (2005) Incremental On-line Learning in High Dimensions. *Neural Computing* 17(12): 2602-2634.

Watkins, C.J. (1989) *Learning from Delayed Rewards*. Ph.D. Thesis, Cambridge University.

Wilson, S.W. (1981) Aubert Processing and Intelligent Vision. Technical Report: Polaroid Corporation.

Wilson, S.W. (1985) Knowledge Growth in an Artificial Animal. In J.J. Grefenstette (ed) *Proceedings of the First International Conference on Genetic Algorithms and their Applications.* Lawrence Erlbaum Associates, pp16-23.

Wilson, S.W. (1987) Classifier Systems and the Animat Problem. *Machine Learning* 2: 219-228.

Wilson, S.W. (1994) ZCS: A Zeroth-level Classifier System. *Evolutionary Computation* 2(1):1-18.

Wilson, S.W. (1995) Classifier Fitness Based on Accuracy. *Evolutionary Computation* 3(2): 149-76.

Wilson, S.W. (1998) Generalization in the XCS Classifier System. In Koza et al. (eds.) *Genetic Programming 1998: Proceedings of the Third Annual Conference,* pp. 322-334. Morgan Kaufmann.

Wilson, S. W. (2002) Classifiers that Approximate Functions. *Natural Computing* 1(1): 211-233.

Wilson, S.W. (2007) Three Architectures for Continuous Action. In J. Bacardit et al. (eds) *Learning Classifier Systems: Revised Selected Papers*. Springer, pp239-257.

Wilson, S.W. (2008) Classifier Conditions using Gene Expression Programming. In J. Bacardit et al. (eds) *Learning Classifier Systems: Revised Selected Papers*. Springer, pp206-217.

Wilson, S.W. & Goldberg, D.E. (1989) A critical review of classifier systems. In J. Schaffer (ed) *Proceedings of the 3rd International Conference on Genetic Algorithms*, Morgan Kauffman, pp244-255.

Wyatt, D. & Bull, L. (2004) A Memetic Learning Classifier System for Describing Continuous-Valued Problem Spaces. In N. Krasnagor, W. Hart & J. Smith (eds) *Recent Advances in Memetic Algorithms*. Springer, pp355-396

Xu, R. & Wunsch, D. (2009) *Clustering*. IEEE Press.